\documentclass[conference]{IEEEtran}
\IEEEoverridecommandlockouts

\usepackage{cite}
\usepackage{amsmath,amssymb,amsfonts}
\usepackage{algorithmic}
\usepackage{graphicx}
\usepackage{textcomp}
\usepackage{xcolor}

\usepackage{url}
\newcommand{\HeI}{\ion{He}{i} 10830 \AA{}} 
\newcommand{\Ha}{H$\alpha$}
\DeclareRobustCommand{\ion}[2]{%
\relax\ifmmode
\ifx\testbx\f@series
{\mathbf{#1\,\mathsc{#2}}}\else
{\mathrm{#1\,\mathsc{#2}}}\fi
\else\textup{#1\,{\mdseries\textsc{#2}}}%
\fi}

\def\BibTeX{{\rm B\kern-.05em{\sc i\kern-.025em b}\kern-.08em
    T\kern-.1667em\lower.7ex\hbox{E}\kern-.125emX}}

\begin{document}

\title{Deep Computer Vision for Solar Physics Big Data: Opportunities and Challenges}

\author{\IEEEauthorblockN{Bo Shen\textsuperscript{1*}\thanks{\textsuperscript{*}Dr. Bo Shen is the corresponding author. Email: bo.shen@njit.edu }, Marco Marena\textsuperscript{1}, Chenyang Li\textsuperscript{1}, Qin Li\textsuperscript{1}, Haodi Jiang\textsuperscript{2}, \\ Mengnan Du\textsuperscript{1}, Jiajun Xu\textsuperscript{3}, Haimin Wang\textsuperscript{1}}
\IEEEauthorblockA{\textsuperscript{1}New Jersey Institute of Technology, Newark, NJ} 
\IEEEauthorblockA{\textsuperscript{2}Sam Houston State University, Huntsville, TX; \textsuperscript{3}Meta, Menlo Park, California} 
}

\maketitle

\begin{abstract} 
With recent missions such as advanced space-based observatories like the Solar Dynamics Observatory (SDO) and Parker Solar Probe, and ground-based telescopes like the Daniel K. Inouye Solar Telescope (DKIST), the volume, velocity, and variety of data have made solar physics enter a transformative era as solar physics big data (SPBD). With the recent advancement of deep computer vision, there are new opportunities in SPBD for tackling problems that were previously unsolvable. However, there are new challenges arising due to the inherent characteristics of SPBD and deep computer vision models. This vision paper presents an overview of the different types of SPBD, explores new opportunities in applying deep computer vision to SPBD, highlights the unique challenges, and outlines several potential future research directions.
\end{abstract}
\begin{IEEEkeywords}
Deep Computer Vision, Solar Physics, Big Data 
\end{IEEEkeywords}

\section{Introduction} \label{sec: intro}
Solar physics, which focuses on studying the Sun and its interactions with the solar system, has always been data-intensive. Recent missions have been launched such as the advanced space-based observatories like the Solar Dynamics Observatory (SDO) \cite{pesnell2012solar} and Parker Solar Probe \cite{case2020solar}, and ground-based telescopes like the Daniel K. Inouye Solar Telescope (DKIST) \cite{rimmele2020daniel}, providing high-resolution images and continuous monitoring of solar activities. Specifically, access to the curated SDO data \cite{galvez2019machine} implies downloading 6.5 terabytes (TB) of data. The estimated amount of data in an excellent observing day for DKIST easily reaches the petabyte (PB) regime \cite{asensio2023machine}. These solar data, characterized by their volume, velocity, and variety, have made solar physics enter a transformative era with the advent of big data, referred to as \textit{solar physics big data} (SPBD) \cite{banda2014big}. 

As an observational science, we observe solar data in solar physics since we cannot change the experimental conditions. Analyzing solar data has always been researchers’ interest, fostering the development of solar physics data mining \cite{battams2014stream},  a research field that aims to discover non-trivial, previously unknown, but potentially useful patterns or make accurate predictions based on solar data. Due to the inherent characteristics of SPBD, the data's complexity and size often exceed the capabilities of traditional methods, including statistical learning techniques. This necessitates developing and applying more efficient methods to manage and analyze SPBD effectively. 


Thanks to advanced cyberinfrastructure, efficient computational hardware (e.g., GPUs), carefully designed neural network architecture, and learning algorithms, \texttt{deep computer vision}, i.e., deep learning-based computer vision, has brought new opportunities to various domains. Specifically, it has achieved tremendous success in solar physics \cite{asensio2023machine} by performing tasks such as image segmentation, super-resolution, image-to-image translation, etc. 

When SPBD meets deep computer vision, significant new opportunities emerge due to the recent increase in the capacity of deep models (i.e., deep neural networks). Deep computer vision has significantly transformed the analysis of SPBD, making it possible to automate and process vast and complex datasets with exceptional accuracy and speed. One of the key transformative aspects is its ability to perform automatic feature extraction/engineering, eliminating the need for manually identifying relevant patterns and structures within solar data. Therefore, these powerful models can now detect and classify intricate solar phenomena, such as sunspots, solar flares, and coronal mass ejections (CMEs), with great precision and in real time. As a result, the capacity of deep computer vision allows for the extraction of hidden patterns within SPBD that traditional models may overlook. 

This vision paper will explore emerging opportunities in applying deep computer vision to solar physics big data, highlight the unique challenges, and outline several potential future research directions.

\section{Types of SPBD}
\subsection{Ground Imaging and Spectroscopy of Many Decades}
Ground-based observations have a long history over many decades. \Ha{} (656.3 nm), \HeI{} (1083 nm), and Ca II K (393.37 nm) are spectral lines for studying different layers of the Sun's atmosphere. \Ha{}, a prominent line in the red visible spectrum, is associated with the chromosphere and is widely used to observe solar prominences, filaments, and flares. \Ha{} images are available from several major observatories like Global Oscillation Network Group (GONG), Big Bear Solar Observatory (BBSO), and Kodaikanal Solar Observatory (KSO). \HeI{}, located in the infrared spectrum, probes the upper chromosphere and lower corona, providing insights into magnetic fields and coronal holes. \HeI{} images can be accessed through the National Solar Observatory (NSO) and Mauna Loa Solar Observatory. Ca II K, found in the near-ultraviolet, focuses on the lower chromosphere and upper photosphere, making it essential for studying plages, chromospheric network structures, and long-term solar activity. Ca II K images are from KSO, BBSO, and the Meudon Solar Observatory, among others.

\subsection{Line-of-sight and Vector Magnetograms}
Magnetograms are specialized images that visualize the magnetic field strength and orientation on the Sun's surface (photosphere). These images are captured by various solar observatories and instruments, each contributing to a comprehensive understanding of solar magnetic activity. The SOHO’s Michelson Doppler Imager (SOHO/MDI) was among the first to provide long-term, consistent magnetogram data, laying the groundwork for solar magnetic field research. 
The Hinode's Solar Optical Telescope (Hinode/SOT) offers detailed magnetograms that help refine our understanding of the Sun’s magnetic field structure and dynamics, especially at smaller scales. The SDO's Helioseismic and Magnetic Imager (SDO/HMI) provides high-resolution, continuous observations of the Sun's magnetic fields.  The BBSO's Goode Solar Telescope (BBSO/GST) also contributes high-resolution magnetograms, focusing on capturing fine details of solar magnetic fields. 

\subsection{In-Space Observations — Extreme Ultraviolet} 
Extreme Ultraviolet (EUV) images captured by the SDO's Atmospheric Imaging Assembly (SDO/AIA), the SOHO's Extreme ultraviolet Imaging Telescope (SOHO/EIT), and the Solar Terrestrial Relations Observatory's Sun Earth Connection Coronal and Heliospheric Investigation (STEREO/SECCHI) provide critical in-space observations of the Sun's corona. These instruments, positioned in space to avoid atmospheric interference, offer a direct and unfiltered view of the Sun's outer atmosphere, where temperatures reach millions of degrees. SDO/AIA provides high-resolution images at multiple wavelengths of EUV light with a rapid cadence, offering near-real-time monitoring of solar activity from its geosynchronous orbit around Earth. SOHO/EIT, positioned at the L1 Lagrange point, has been providing long-term EUV data since 1996, contributing to our understanding of solar cycles and long-term solar activity. STEREO/SECCHI, with its two spacecraft in different orbits around the Sun, offers unique stereoscopic views, enabling three-dimensional analysis of solar events like CMEs as they travel through space. 





\section{New opportunities of deep computer vision for solar physics big data}
\subsection{Why Deep Computer Vision for SPBD?}

First, deep computer vision models, characterized by their extensive number of parameters (usually more than millions) are highly effective at handling large volumes of complex data. These models can detect complex solar patterns effectively, which is crucial for studying intricate physical processes that involve various interacting variables across diverse spatial and temporal domains. 

Second, deep computer vision enables end-to-end training that eliminates the need for manual feature engineering. Traditional solar physics data mining requires feature engineering with domain-specific knowledge, which is labor-intensive. This process must be repeated for each new application, which is both time-consuming and potentially inadequate for capturing intricate patterns.  Deep computer vision can directly learn features/patterns from the data, minimizing the effort and time required for feature engineering. For instance, researchers have trained deep models for solar flare prediction using full-disk line-of-sight magnetograms provided by SDO/HMI \cite{pandey2021deep,pandey2021solar,pandey2022towards}, without manually extracting physics-based features.

Third, Large language model (LLM) has been very successful recently. It also has been applied to the science field \cite{taylor2022galactica} already. SPBD involves multiple data modalities, such as images and text data. Therefore, SPBD is well suited to apply Vision Language Models (VLMs) to analyze different types of data at once, where deep computer vision is one important component in VLMs. VLMs have demonstrated that with sufficient training data, it’s possible to train a general model capable of performing well across diverse data sources.   

\subsection{Major Transformative Applications}
\subsubsection{Image Segmentation}
Deep computer vision has significantly advanced image segmentation of solar images, which is particularly important given the vast quantity of solar images/videos available every day. Automating the detection and segmentation of solar structures within these images could enable the creation of extensive databases. Sunspots, flares, coronal holes, and other solar features are prime candidates for these applications. Illarionov and Tlatov \cite{illarionov2018segmentation} applied U-Net architecture to identify coronal holes on SDO/AIA images. Jiang et. al \cite{jiang2020identifying} proposed SolarUnet to identify and track solar magnetic flux elements or features in observed vector magnetograms based on the Southwest Automatic Magnetic Identification Suite. Later, Jiang et. al \cite{jiang2021tracing} proposed FibrilNet for tracing chromospheric fibrils in \Ha{} images of solar observations from BBSO/GST. Castillo et. al \cite{diaz2022towards} presented the first attempts to classify and identify structures in the solar granulation based on the U-Net architecture.  Zheng et. al \cite{zheng2024developing} utilized the U-Net to identify filaments and implement the Channel and Spatial Reliability Tracking algorithm for automated filament tracking of CHASE/HIS \Ha{} images.


\subsubsection{Super-resolution}
Super-resolution (SR) is a widely used technique that aims to improve the spatial resolution of images. In solar physics, observational instruments like telescopes and imaging satellites have inherent resolution limits based on their design, such as the size of the optical aperture and the quality of the detectors. 
Recent studies in solar physics have utilized advanced SR methods to improve the quality of solar images. Researchers have improved the spatial resolution of SDO/HMI images to match the spatial resolution of BBSO/GST.  Deng et al. \cite{deng2021improving} used a combination of generative adversarial networks (GAN) \cite{goodfellow2014generative} and self-attention mechanisms. Song et al. \cite{song2022improving} proposed an improved conditional denoising diffusion probability model (ICDDPM). 
Researchers also tried to improve the spatial resolution of magnetograms. Baso and Ramos \cite{baso2018enhancing} developed a method called Enhance to improve the resolution of SDO/HMI magnetograms to the high-resolution image from Hinode. 
Xu et al. \cite{xu2024super} developed SolarCNN, a residual attention-aided convolutional neural network to improve the resolution of SOHO/MDI magnetograms using SDO/HMI data as high-resolution ground truth.


\subsubsection{Image-to-image Translation}
Image-to-image translation is a powerful technique in computer vision that involves converting images from one domain to another. The pix2pix, developed by Isola et al. \cite{isola2017image}, is one of the most influential tools in this domain, which has been extensively applied in solar physics. 
One application is cloud removal in \Ha{} images. Wu et al. \cite{wu2022algorithm} employed a pix2pix network with a U-Net generator and PatchGAN discriminator to effectively remove cloud shadows.  
Ma et al. \cite{ma2024cloud} introduced RPix2PixHD, which integrates a pix2pixHD network with a registration network to handle misaligned image pairs. 
Another critical application is generating synthetic solar images.
Park et al. \cite{Park_2019} used pix2pixHD to create synthetic ultraviolet (UV) and EUV solar images from SDO/HMI images. 
Son et al. \cite{Son_2021} generated \HeI{} images from SDO/AIA images, successfully producing high-resolution synthetic images that closely match real observations, thus filling gaps caused by atmospheric conditions or equipment issues.


\subsubsection{Stokes Inversion}
Obtaining high-quality magnetic and velocity fields through Stokes inversion is crucial in solar physics \cite{jiang2022inferring}, as the Sun's magnetic field drives all solar activities. 
Observational instruments do not directly measure the magnetic field; instead, Stokes inversion is used to infer the physical conditions of the solar atmosphere by interpreting observed Stokes profiles. Various inversion methods, such as the Milne–Eddington (ME) approximation, aim to achieve the best fit to the observed Stokes profiles. Notable implementations include SPINOR, Helix+, and VFISV \cite{liu2020inferring}. However, traditional Stokes inversion methods are time-consuming. The advent of powerful telescopes like the DKIST and BBSO/GST, which generate large volumes of solar Stokes profile data daily, exacerbates these challenges, pushing the limits of these technologies \cite{yang2024spectropolarimetric}. 

Efforts are being made to employ deep learning techniques to accelerate Stokes inversion within a practical timeframe. Asensio Ramos and Díaz Baso \cite{ramos2019stokes} presented two convolutional neural networks for Stokes inversion on synthetic 2D maps, using spatial coherence to produce a 3D cube of thermodynamic and magnetic properties. Liu et al.\cite{liu2020inferring} designed a pixel-level CNN (PCNN) to perform Stokes inversion on GST/NIRIS data at BBSO, producing magnetic field vectors from Stokes $Q$, $U$, and $V$ profiles. Later, Jiang et al. developed a stacked Deep Neural Network to invert all four Stokes profiles from GST/NIRIS, producing magnetic and velocity field vectors \cite{jiang2022inferring}. 
Higgins et al. \cite{higgins2021fast} employed a U-Net CNN to speed up vector magnetogram production from SDO/HMI, refining it with Hinode/SOT-SP data, leading to SynthIA's development \cite{higgins2022synthia}. Yang et al. recently introduced the SPIn4D project to develop CNNs to rapidly estimate solar photosphere properties from DKIST observations using large-scale MHD simulations and synthetic Stokes profiles \cite{yang2024spectropolarimetric}.


\section{Unique Challenges}
Despite the recent development of deep computer vision for SPBD, there are several unique challenges as below.
\subsection{Image Cleaning, Registration, Alignment}
The integration and alignment of multi-instrument data, such as \Ha{} from BBSO and magnetograms from instruments like SDO, require image cleaning, registration, and alignment methods to ensure consistency and accuracy across different data sources. In-space instruments often benefit from more stable observing conditions, while ground-based telescopes are affected by atmospheric distortions such as turbulence, absorption, and scattering, which can degrade data quality. These atmospheric effects needed to be corrected when pairing ground-based data with in-space observations before registration/alignment. Challenges also include variations in spatial and temporal resolution, observational perspectives, and differences in wavelength bands and filters by different instruments. For example, the Sun's rotation and rapidly changing features would change variations in spatial and temporal resolution. Differences in the field of view between instruments pose additional challenges when pairing data, particularly when comparing large-scale features observed from space with fine-scale structures captured from the ground. 

\subsection{Reliability of AI-generated Data}

One of the most significant challenges in applying deep computer vision techniques to SPBD is ensuring that AI-generated data and predictions are not only statistically accurate but also meaningful in physics. Domain-specific knowledge and physics-based evaluations are crucial for solar physics applications. For instance, when applying super-resolution techniques to magnetograms, the need for physics evaluation approaches is highlighted in \cite{2024ApJS..271...46M}, such as verifying the integration of radial magnetic field components (assuming the fields are divergence-free) and gradient evaluation to examine polarity inversion lines between bipolar magnetic field regions. These physics-based metrics guide the design of loss functions to maintain the physical integrity of the synthetic data. 
Researchers also applied AI-generated data to real applications as an evaluation approach. 
 AI-generated synthetic magnetograms are used to address data imbalance issues in flare prediction, achieving higher prediction accuracy \cite{2024ApJS..271...29A}. Similarly, temporal profiles of magnetic parameters are aligned with flare eruptions to validate the reasonableness of magnetic field extrapolation results \cite{2023NatAs...7.1171J}. 

\subsection{Black Box of Deep Computer Vision}
In solar physics, there are well-developed physics models such as magnetohydrodynamic modeling \cite{gombosi2018extended}. Compared to these physics-based models, deep neural networks are often considered black boxes with limited explainability, which raises concerns, especially when the goal is to understand and advance physics. The opacity of these deep models makes it challenging to trace and explain how they transform inputs into outputs, complicating debugging, refining, and determining when the model might fail. For example, when we use SDO/HMI images for solar flare prediction, deep models cannot tell which part of the image is important for flare prediction. Addressing these challenges through research in interpretable and explainable AI is crucial for the responsible and widespread use of deep computer vision.

\section{Future Research Directions}
\subsection{AI/ML-ready Dataset} 
Solar physicists have begun to use AI/ML to achieve breakthroughs. The AI/ML approach is data-intensive, where datasets are used for training, tuning, and testing AI/ML. It starts with creating ``clean” datasets which require fixing structural errors, handling missing data, removing non-physical outlier points, and/or filtering observations. With a clean dataset, even simple algorithms can yield important insights. The data quality is key to the performance of AI/ML algorithms. Therefore, solar physics data must be AI/ML-ready to apply various methods and tools and to be stored as AI/ML catalogs and archives for public use.  Traditional methods of data preparation/creation are often labor-intensive and time-consuming, limiting the speed and efficiency of research. An automated pipeline such as the ones in \cite{kucuk2017large,angryk2020multivariate} is needed to streamline the process of data acquisition, cleaning, normalization, and annotation, ensuring that datasets are not only accurate but also optimized for developing AI/ML models. Such a system would significantly enhance the capability to handle large-scale solar datasets, allowing researchers to focus on developing and refining models rather than on preliminary research on a small-scale dataset.


\subsection{Physics-informed Computer Vision} Most deep computer vision models in SPBD are purely data-driven, and their results can not be considered fully reliable for solar physicists. These data-driven models often lack the underlying physical understanding necessary to accurately analyze and interpret solar phenomena, which are governed by intricate physical processes.  To address this issue, physics-informed computer vision is crucial in solar physics research because it allows for the integration of established physical laws and domain-specific knowledge directly into deep computer vision models. For example, physics-informed neural networks (PINNs) have been applied to accelerate solar coronal magnetic fields modeling \cite{jarolim2023probing,baty2024modelling}. Physics-informed computer vision combines data with known underlying physical laws, making them more effective than models that rely only on data. In solar physics, this approach helps the models make better predictions, even when there's limited data available. It also ensures that the predictions make sense according to the laws of physics, which makes the results more trustworthy and easier to understand. This synergy between physics and AI/ML ultimately accelerates scientific discovery and improves our ability to mitigate space weather risks. Without incorporating domain-specific knowledge, data-driven methods can lead to overfitting, especially when faced with limited or noisy data. %

\subsection{Interpretability and Explainability}
Apart from physics-informed computer vision, interpretable and explainable AI is essential to open the black box of deep models in solar physics research because it helps researchers understand how deep models make predictions. Solar physics involves studying complex phenomena like solar flares, sunspots, and magnetic fields, which are difficult to predict and analyze. When deep computer vision models are used to study these events, it's important that scientists can trace the model's decisions back to the specific data or features that influenced them. This transparency allows scientists to check whether the predictions make sense based on what is already known about the Sun and can even help uncover new patterns or insights that were previously hidden. By making deep models more interpretable, scientists can trust the results more, collaborate better with AI experts, and use these tools more effectively to push the boundaries of our understanding of solar activity.  For example, researchers have been working on the interpretability and explainability of solar flare prediction models \cite{sun2021improved,pandey2023explainable,pandey2023explaining}.


\subsection{Vision Language Model}
VLMs such as Minigpt-4 \cite{zhu2023minigpt} could revolutionize the field by enabling a more comprehensive analysis of solar events.  These advanced models can seamlessly integrate visual data from solar images with descriptive textual information, significantly enhancing the depth and precision of analyzing complex solar activities and events.
For instance, by combining high-resolution images of solar flares, sunspots, and CMEs with detailed contextual explanations, VLMs could play a crucial role in automating the classification of various solar features. Additionally, these advanced models can detect patterns that might otherwise go unnoticed and predict solar events with improved accuracy. The fusion of visual and text data helps us make more insightful discoveries and improves our understanding of how the Sun works. This, in turn, advances forecasting capabilities, ultimately contributing to more reliable space weather predictions, which are vital for protecting both terrestrial and space-based technological systems.

\bibliographystyle{IEEEtran}
\bibliography{IEEEfull,Example}

\end{document}